%% file: SVGAN.tex
\documentclass[letterpaper,twocolumn,fleqn]{article} 

\usepackage{ist}
\usepackage{amssymb}
\usepackage{color}
\usepackage{algorithm}
\usepackage{algorithmic}
\usepackage{array}
\usepackage{tikz}
\usetikzlibrary{shapes}
\usepackage{amsmath}
\usepackage{url}
\usepackage{graphicx}
\usepackage{tabularx}
\usepackage[stable]{footmisc}
\usepackage{xcolor}
\usepackage{caption}
\usepackage{hyperref}
 
\def\##1{\relax\ifmmode\mathchoice
{\mbox{\boldmath$\displaystyle#1$}}
{\mbox{\boldmath$\textstyle#1$}}
{\mbox{\boldmath$\scriptstyle#1$}}
{\mbox{\boldmath$\scriptscriptstyle#1$}}\else
\hbox{\boldmath$\textstyle$}\fi}

\title{Yes, we GAN: Applying Adversarial Techniques for Autonomous Driving}

\author{Michal U\v{r}i\v{c}\'{a}\v{r}$^{1}$, Pavel K\v{r}\'{i}\v{z}ek$^{1}$, David Hurych$^{1}$, Ibrahim Sobh$^{2}$, Senthil Yogamani$^{3}$ and Patrick Denny$^{3}$ \\
$^{1}$Valeo DVS Prague, Czech Republic \\
$^{2}$Valeo CDV AI Cairo, Egypt \\
$^{3}$Valeo Vision Systems, Ireland
}

\begin{document} 

\maketitle 



\input{include/abstract.tex}


\input{include/introduction.tex}

\input{include/overviewgans.tex}

\input{include/ganapplicationsforad.tex}

\input{include/results.tex}

\input{include/discussion.tex}

\input{include/conclusions.tex}

\section*{ACKNOWLEDGMENTS} 
The authors would like to thank their employer for the opportunity to perform fundamental research. We would also like to thank the company Borealis AI for the GAN T-shirts which inspired the title of this paper\footnote{https://www.borealisai.com/en/blog/yes-we-gan/}. Last but not least, we would like to express our very special thanks to our colleague GANesh Sistu, who gave us permission to use his nice figures.



{\small
\bibliographystyle{ieeetr}
\bibliography{references}
}


\input{include/aboutauthors.tex}

\end{document}

%% file: include/abstract.tex
\begin{abstract}
Generative Adversarial Networks (GAN) have gained a lot of popularity from their introduction in 2014 till present. Research on GAN is rapidly growing and there are many variants of the original GAN focusing on various aspects of deep learning. GAN are perceived as the most impactful direction of machine learning in the last decade. This paper focuses on the application of GAN in autonomous driving including topics such as advanced data augmentation, loss function learning, semi-supervised learning, etc. We formalize and review key applications of adversarial techniques and discuss  challenges and open problems to be addressed.
\end{abstract}

%% file: include/introduction.tex
\section{INTRODUCTION}\label{sec:introduction} 

\begin{figure*}[tb]
\centering
\includegraphics[width=0.95\linewidth]{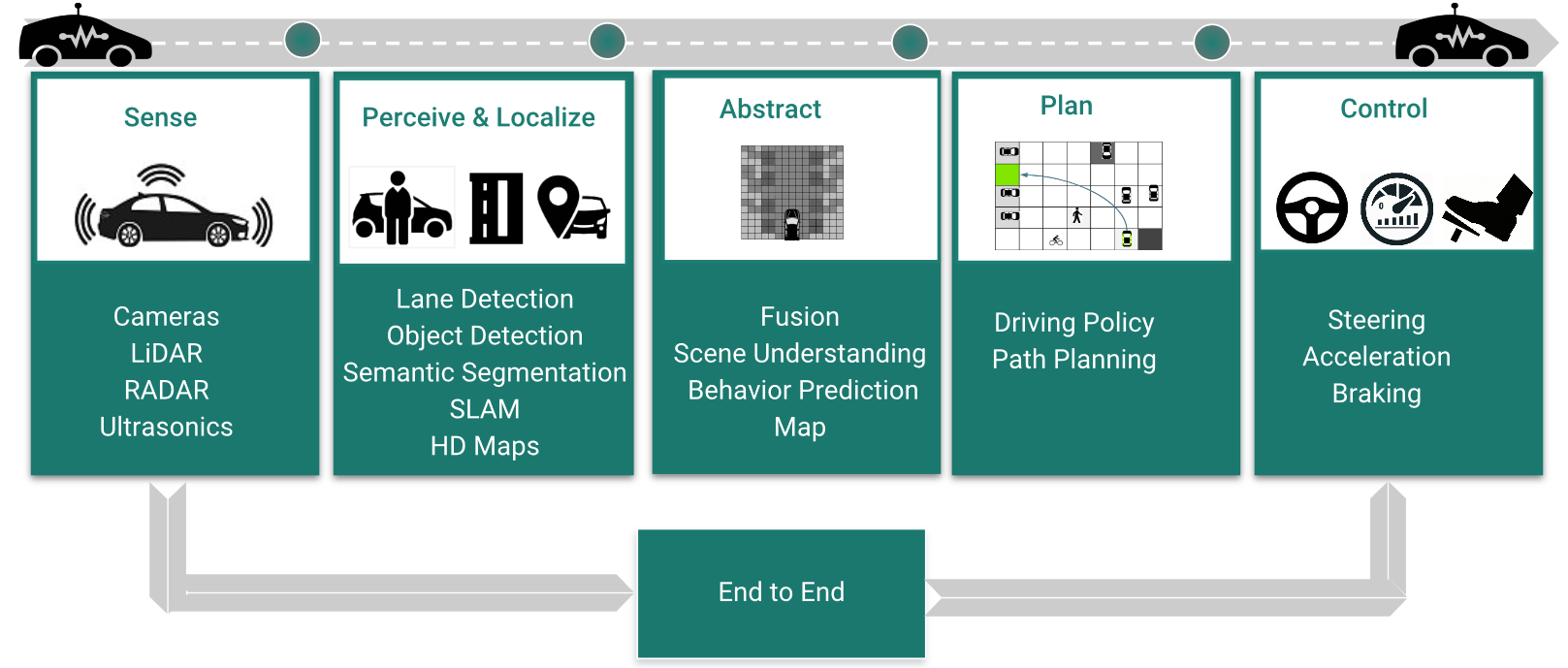}
\caption{Modules in an autonomous driving pipeline. 
}
\label{fig:ADBlockDiagram}
\end{figure*}

Autonomous driving is becoming a common feature in modern vehicles. Fully autonomous driving still remains a challenging task and there is a lot of active research to solve technical problems. Figure \ref{fig:ADBlockDiagram} illustrates the standard modules in an autonomous driving system.  The first stage is perception using a suite of sensors. Common sensors which are already deployed are Cameras, Radar, Ultrasonics and Lidar. The first stage in the processing is perception where semantic objects like lanes/vehicles or geometric objects like freespace and generic obstacles are detected. They are then fused into a generic abstract representation typically a 2D or 3D map of objects with respect to the ego vehicle. A driving policy algorithm uses this map to decide a trajectory for the car to maneuver. The traditional approach is to have the modules independently designed but there are also attempts to do end to end learning.

\begin{figure}[tb]
\centering
\includegraphics[width=\linewidth]{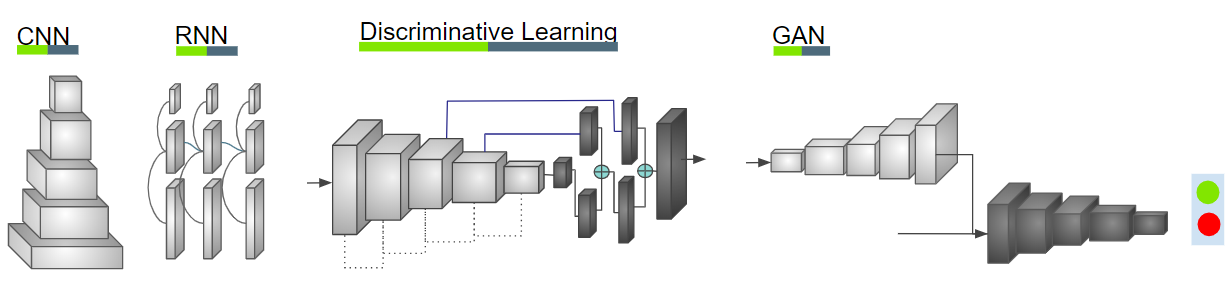}
\caption{Commonly used deep learning models - CNN, RNN and GAN. 
}
\label{fig:cnn_gan}
\end{figure}

Progress in Deep learning has accelerated the maturity of autonomous driving systems \cite{siam2017deep}. Deep learning models have become a standard in perception and gradually becoming competitive for other modules like fusion and driving policy. As shown in Figure \ref{fig:cnn_gan}, the commonly used deep learning models are Convolutional Neural Networks (CNN) and Recurrent Neural Networks (RNN). These are discriminative models which are trained for classification or regression problems. Discriminative models extract features which are sufficient to solve the classification problem and do not typically capture the complete information in the data. Generative models on the other hand try to capture the data distribution and hence form a more powerful representation. Generative Adversarial Networks (GAN) belong to this category and has become an effective generative model in various domains and tasks. 

GAN have been progressing very rapidly and are seen as one of the most impactful models in the field of machine learning. The realism of generation of images using GAN has been impressive and recent results can capture subtle expressions. In spite of its popularity in the field of machine learning, it is relatively less explored for applications of autonomous driving. The main application where GAN are used in AD are image-to-image translation for style transfer from synthetic to realistic or transfer across different conditions of lighting, weather, etc. Autonomous driving systems have to be extremely robust and this requires training the model with all possible scenarios which can happen in real life. Collecting such a dataset is in-feasible in practice and synthetic data simulators are commonly used to ameliorate this issue. Generative models, like GAN, provide a promising path to generate realistic datasets in this case. 

The motivation of this paper is to survey adversarial techniques, distill the key ideas for applications of automated driving and provide a list of challenges and open problems from our industrial experience. The rest of the paper is organized into four sections. The first section describes the vanilla GAN as proposed in the original paper~\cite{GAN}, prominent derivatives and recent advance.  The second section discusses applications of GAN for autonomous driving. The third section discusses the main applications of GAN in autonomous driving. The fourth section summarizes results of our experiments on using GAN for soiling and adverse weather classification. The fifth section provides the discussion of the main challenges arising with GAN. Finally, the last section summarizes the paper and provides concluding remarks.

%% file: include/overviewgans.tex
\section{OVERVIEW OF GAN} 
\label{sec:gan}


Generative Adversarial Networks were introduced in 2014 and were immediately recognized as a perspective direction of upcoming deep learning research, especially in domains such as unsupervised and semi-supervised learning, or advanced data augmentation.
The idea behind GAN is very simple and intuitive, which might be one of the reasons for such a big popularity in the research community. Moreover, it seems that GAN are attracting interest of not only research pioneers, but as well the proof-seeking theoreticians, since both branches are getting more or less equal attention, and there appear both papers about new applications as well as theoretical improvements of the existing drawbacks of GAN. Therefore, we can see interesting applications such as unsupervised image-to-image translation, where for example images with horses are successfully transformed to realistically looking images with zebras, or a shot of Yosemite National Park taken in summer, transformed to the same scenery how it would probably look like in winter~\cite{Isola-CVPR-2017,Zhu-ICCV-2017}. The theory-deepening papers on the other hand deal with the stabilization of the complicated GAN learning, which often tends to get stuck in the mode collapse, i.e. the state when the discriminator is fooled to believe in unrealistic samples or degeneration of the generator, providing only a very limited set of samples~\cite{salimans2016improved,WassersteinGAN,Gulrajani-NIPS-2017}. The mode collapse problem arises when the generated data do not reflect the diversity of the underlying data distribution.


\subsection{Vanilla GAN}

\begin{figure}[tb]
    \includegraphics[width=0.95\linewidth]{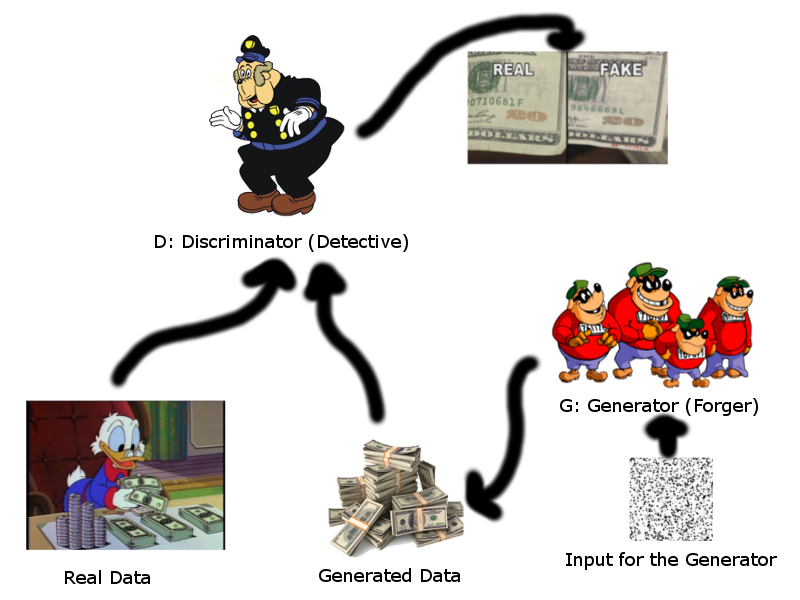}
    \caption{The conceptual idea of Generative Adversarial Networks. Generator (Forger) takes random noise on input and generates fake data. Discriminator (Detective) takes samples from both real and generated data and is trying to distinguish both categories, by reporting real or fake output for each sample.}
    \label{fig:GAN_concept}
\end{figure}

Vanilla GAN~\cite{GAN} were introduced as a two-player minimax game, where each player is represented by a neural network. One network is a generator and the other one is a discriminator. The generator's task is to generate samples, which are as similar to the real data samples as possible, while the discriminator's task is to distinguish the real samples from the generated ones. See Figure~\ref{fig:GAN_concept} for the concept idea. The optimization task should end up at an equilibrium point, where the generator is able to generate samples, which the discriminator cannot distinguish from the real ones. In other words, the discriminator should output a probability equal to $0.5$ for either of the real or generated input. 

By using the neural networks, one of the biggest problems connected to generative modeling is mitigated--- generative modeling in computer vision problems usually requires very complicated sampling functions, or complicated structures, and often only approximate inference computation is possible. However, using the neural networks is very simple, gradients are computed by a simple, yet effective, back-propagation algorithm, and the intuition behind GAN is that the complicated sampling function can be constructed by implicit learning.

Let us describe GAN more formally. We assume for a simplicity that both models (generator and discriminator) are multi-layer perceptrons. The task is to learn the generator's distribution $p_g$, using the data samples $\#x$. Then, let us define a prior on the input noise variables $p_{\#z}(\#z)$. Using these, we can define a mapping to the data space as a differentiable function~$G$, represented by a multi-layer perceptron with parameters $\theta_g$: $G(\#z;\ \theta_g)$. Function $G$ represents our generator. Next, we define a second function $D$, also as a multi-layer perceptron, with parameters $\theta_d$: $D(\#x;\ \theta_d)$, the goal of which is to represent a probability that $\#x$ comes from the data, and not from $p_g$. The function $D$ represents our discriminator. 

$D$ is trained to maximize the probability of assigning the correct label to both the training samples and samples generated by $G$. Simultaneously, $G$ is trained to minimize the $\log(1 - D(G(\#z))$. This refers to what we have already mentioned in the beginning--- $D$ and $G$ are playing a two-player minimax game defined as follows: 
\begin{eqnarray}
	\min_{G}\max_{D} V(D, G) \!\!\!\!\! &=& \!\!\!\!\! \mathbb{E}_{\#x \sim p_{\mathrm{data}(\#x)}} \big[ \log D(\#x) \big] \nonumber \\ 
    \!\!\!\!\! & & \!\!\!\!\! + \ \mathbb{E}_{\#z \sim p_{\#z}(\#z)} \Big[ \log\big(1 - D(G(\#z))\big) \Big] \;, \nonumber \\ 
\end{eqnarray}
where $V(G, D)$ is a value function. 

Because of the computational complexity of the discriminator maximization, the game has to be solved by an iterative numerical approach, which is summarized in Algorithm~\ref{alg:minibatch_SGD}.

\begin{algorithm}[tbh]
	\centering
    \caption{Mini-batch Stochastic Gradient Descent (SGD) training of GAN. The number of steps to apply to the discriminator, $k$, is a hyper-parameter. In the original paper~\cite{GAN}, the authors used $k=1$.}
    \label{alg:minibatch_SGD}
    \begin{algorithmic}[1]
    	\FOR {number of training iterations}
            \FOR {$k$ steps}
            	\STATE Sample mini-batch of $m$ noise samples $\{\#z^{(1)}, \dots, \#z^{(m)} \}$ from noise prior $p_{g}(\#z)$.
                \STATE Sample mini-batch of $m$ samples $\{\#x^{(1)}, \dots, \#x^{(m)} \}$ from data generating distribution $p_{\mathrm{data}}(\#x)$.
                \STATE Update the discriminator by ascending its stochastic gradient:
                \begin{equation}
                	\nabla_{\theta_d} \frac{1}{m} \sum_{i=1}^{m} \left[ \log D \left( \#x^{(i)} \right) \!+\! \log \left( 1 \!\!-\!\! D \left( G \left( \#z^{(i)} \right) \right) \right) \right]. \nonumber
               	\end{equation}
        	\ENDFOR
            \STATE Sample mini-batch of $m$ noise samples $\{\#z^{(1)}, \dots, \#z^{(m)} \}$ from noise prior $p_{g}(\#z)$.
            \STATE Update the generator by descending its stochastic gradient:
            \begin{equation}
            	\nabla_{\theta_g} \frac{1}{m} \sum_{i=1}^{m} \log \left( 1 - D \left( G \left( \#z^{(i)} \right) \right) \right). \nonumber
            \end{equation}
        \ENDFOR
    \end{algorithmic}
\end{algorithm}

We will here omit the theoretical results, such as the global optimality of $p_g = p_{\mathrm{data}}$, or the proof of convergence of the Algorithm~\ref{alg:minibatch_SGD}, those are described in the original paper~\cite{GAN} and are not particularly interesting for the potential applications of the GAN in various tasks. Instead, we will just summarize the advantages and disadvantages of GAN in the next paragraphs. 

The main disadvantage of GAN framework is the absence of explicit representation of $p_g(\#x)$, and the requirement of simultaneous optimization of the discriminator $D$ with the generator $G$. The authors say explicitly that ``$G$ must not be trained too much without updating $D$''. 

There are several advantages. One was already mentioned--- no inference computation is needed, only back-propagation is needed to compute the gradients. Moreover, a wide variety of functions can be incorporated in the model. 
The comparison to different approaches of generative modeling is summarized in~\cite[Table 2]{GAN}.
Another advantage is that GAN, in contrast to methods based on Markov Chains, can represent very sharp (even degenerate) distributions.

More information about the Vanilla GAN can be found in~\cite{GAN_tutor_Goodfellow-2017}, which summarizes the NIPS 2016 tutorial on GAN.

\subsection{Prominent Derivatives of GAN} \label{sec:GAN_derivatives}

In this section, we outline several prominent derivatives of GAN. While the list of all derivatives is nowadays containing tens of different instances, we will aim only at a few of them. For a reader's convenience, we also list the references in the Table~\ref{tab:gan_derivatives}.

\paragraph*{Conditional Generative Adversarial Nets (CGAN)} were introduced in~\cite{CGAN}. They are basically a straightforward extension of the Vanilla GAN~\cite{GAN}, whereby conditioning the model on additional information one can influence the data generation process. This is crucial for example in the unsupervised learning scenario since we can exploit the conditioning to generate samples of a required label. In other words, we have a possibility to omit the costly annotation process.

The objective function of CGAN is updated by the conditional variable $\#y$ (e.g. a class label) as follows:
\begin{eqnarray}
	\min_{G}\max_{D} V(D, G) \!\!\!\! &=& \!\!\!\! \mathbb{E}_{\#x \sim p_{\mathrm{data}}(\#x)} \left[ \log D(\#x|\#y) \right] \nonumber \\ 
 \!\!\!\!  & & \!\!\!\! + \mathbb{E}_{\#z \sim p_{\#z}(\#z)} \left[ \log \left( 1 - D\left(G\left(\#z|\#y\right)|\#y\right) \right) \right] \;. \nonumber \\
 \label{eq:CGAN}
\end{eqnarray}

CGAN were shown to successfully generate MNIST~\cite{MNIST} digits conditioned on class labels. However, the concept of CGAN was successfully used on more advanced tasks as well, e.g. the fast-converging conditional GAN (FC-GAN)~\cite{FC-GAN} presents results on CIFAR10 dataset~\cite{CIFAR10} as well.

\paragraph*{Wasserstein GAN (WGAN)} is another important derivative of GAN, they were first described in~\cite{WassersteinGAN}. While most of the existing derivatives of GAN are trying to come up with novel or more interesting problems to solve, WGAN is pursuing a different goal--- they focus solely on the learning of GAN. 

In order to learn a probability distribution, one usually use the maximum likelihood estimation (MLE) over the training data. Let us denote the real data distribution (density) as $P_r$, and the parametrized density as $P_\theta$. The MLE amounts to minimization of the Kullback-Leibler divergence $\mathrm{KL}(P_r \parallel P_\theta)$. This requires the model density $P_\theta$ to exist, this requirement is often broken. When we are dealing with distributions supported by low dimensional manifolds, the KL divergence is either not defined or is simply infinite.

The authors of~\cite{WassersteinGAN} provide a theoretical analysis how different distances (such as total variation, KL divergence, Jensen-Shannon divergence, and earth-mover distance, a.k.a. Wasserstein distance) behave in the context of learning distributions. They show on a simple example (which, however, holds for the low dimensional manifolds as well) that except for the Wasserstein distance all other distances are not continuous, and therefore, do not provide clearly defined gradient everywhere. Moreover, Wasserstein distance, besides the guarantees of continuity and differentiability, is also the weakest\footnote{We say that a distance $d$ is weaker than distance $d'$ if every sequence that converges under $d'$ converges also under $d$.} of the distances. There is also a proof that a small Wasserstein distance corresponds to a small distance in distributions. 

The goal of WGAN is, therefore, to minimize the Wasserstein distance $W(P_r, P_\theta)$:
\begin{equation}
	W(P_r, P_\theta) = \inf_{\gamma \in \prod(P_r, P_\theta)}  \mathbb{E}_{(x, y) \sim \gamma} \left[ \left\| x - y \right\| \right] \;. \label{eq:wasserstein1}
\end{equation}
However, the infimum in~(\ref{eq:wasserstein1}) is highly intractable, so the Kantorovich-Rubinstein duality is used, which transforms~(\ref{eq:wasserstein1}) into
\begin{equation}
	W(P_r, P_\theta) = \sup_{\|f\|_{L} \le 1} \mathbb{E}_{x \sim P_r} \left[ f\left(x\right) \right] - \mathbb{E}_{x \sim P_\theta} \left[ f\left(x\right) \right] \;, \label{eq:wasserstein2}
\end{equation}
where the supremum is taken over all $1$-Lipschitz functions. The $K$-Lipschitz function is defined as follows: let $d_X$ and $d_Y$ be distance functions on spaces $X$ and $Y$, respectively. Then a function $f \colon X \rightarrow Y$ is a $K$-Lipschitz if for all $x_1, x_2 \in X$,
\begin{equation}
	d_Y\left( f\left(x_1\right), f\left(x_2\right) \right) \le K d_X(x_1, x_2) \;.
\end{equation}
The gradient of~(\ref{eq:wasserstein2}) is defined as 
\begin{equation}
	\nabla_\theta W(P_r, P_\theta) = -\mathbb{E}_{z \sim p(z)} \left[ \nabla_\theta f\left(g_\theta\left(z\right) \right) \right] \;.
\end{equation}
Finding $f$ maximizing~(\ref{eq:wasserstein2}) is still intractable. We can, however, more easily approximate the problem--- we can use a similar trick as in GAN--- we can train a neural network parametrized with weights $w$ lying in a compact space $\mathcal{W}$ and then backpropagate through $\mathbb{E}_{z \sim p(z)} \left[ \nabla_\theta f\left(g_\theta\left(z\right) \right) \right]$. In order to have parameters $w$ lying in a compact space, we can clamp the weights to a fixed box (e.g. $\mathcal{W} = [-0.01, 0.01]^l$) after each gradient update. The WGAN learning procedure is described in Algorithm~\ref{alg:WGAN}.

\begin{algorithm}
	\centering
    \caption{WGAN algorithm. Authors of~\cite{WassersteinGAN} propose to use the following default values: $\alpha = 0.00005, c = 0.01, m = 64, n_{\mathrm{critic}} = 5$.}
    \label{alg:WGAN}
    \begin{algorithmic}[1]
    \REQUIRE $\alpha$, the learning rate; $c$, the clipping parameter; $m$, the batch size; $n_{\mathrm{critic}}$, the number of iterations of the critic per generator iteration.
    \REQUIRE $w_0$, initial critic parameters; $\theta_0$, initial generator's parameters.
    \WHILE{$\theta$ has not converged} 
    	\FOR {$t = 0, \dots, n_{\mathrm{critic}}$}
        	\STATE Sample $\{ x^{(i)} \}_{i=1}^{m} \sim P_r$ a batch from the real data.
            \STATE Sample $\{ z^{(i)} \}_{i=1}^{m} \sim p(z)$ a batch of prior samples.
            \STATE $g_w \leftarrow \nabla_w \left[ \frac{1}{m} \sum_{i=1}^{m} f_w(x^{(i)}) - \frac{1}{m} \sum_{i=1}^{m} f_w(g_\theta(z^{(i)})) \right]$
            \STATE $w \leftarrow w  + \alpha \cdot \mathrm{RMSProp}(w, g_w)$
            \STATE $w \leftarrow \mathrm{clip}(w, -c, c)$
        \ENDFOR
        \STATE Sample $\{ z^{(i)} \}_{i=1}^{m} \sim p(z)$ a batch of prior samples.
        \STATE $g_\theta \leftarrow -\nabla_\theta \frac{1}{m} \sum_{i=1}^{m} f_w(g_\theta(z^{(i)})) $
        \STATE $\theta \leftarrow \theta - \alpha \cdot \mathrm{RMSProp}(\theta, g_\theta) $
    \ENDWHILE
    \end{algorithmic}
\end{algorithm}

The biggest advantage over vanilla GAN is that we can train the critic\footnote{Critic is a novel name for the discriminator from GAN. The reason for this renaming is, that unlike in GAN, in WGAN the output of this network is not a probability.} till the optimality. This helps to prevent collapsing modes, which is a behavior frequently reported in vanilla GAN.

\paragraph*{Improved WGAN} introduced in~\cite{Gulrajani-NIPS-2017} noticed that the quick and dirty solution of the $1$-Lipschitzness by weight clipping can still lead to flawed learning, where only poor quality samples are generated, or where still convergence failure happens. In~
\cite{Gulrajani-NIPS-2017} a better solution is provided--- the authors propose to penalize the norm of the gradient of the critic with respect to its input and prove that optimal $1$-Lipschitz function's gradient for the optimization criterion in WGAN should have unit norm almost everywhere under $P_r$, and $P_\theta$.

We list the improved WGAN in Algorithm~\ref{alg:iWGAN}, so the reader can compare both variants comfortably. 

\begin{algorithm}
	\centering
    \caption{Improved WGAN algorithm. Authors of~\cite{Gulrajani-NIPS-2017} propose to use the following default values: $\lambda = 10, \alpha = 0.0001, \beta_1 = 0, \beta_2 = 0.9, n_{\mathrm{critic}} = 5$.}
    \label{alg:iWGAN}
    \begin{algorithmic}[1]
    \REQUIRE The gradient penalty coefficient $\lambda$, the number of critic iterations per generator iteration $n_{\mathrm{critic}}$, the batch size $m$, Adam~\cite{Kingma-ICLR-2014} hyperparameters $\alpha,\beta_1,\beta_2$.
    \REQUIRE Initial critic parameters $w_0$, initial generator's parameters $\theta_0$.
    \WHILE{$\theta$ has not converged} 
    	\FOR {$t = 0, \dots, n_{\mathrm{critic}}$}
    	    \FOR {$i = 0, \dots, m$}
    	        \STATE Sample $\{ x^{(i)} \}_{i=1}^{m} \sim P_r$ a batch from the real data; $\{ \#z^{(i)} \}_{i=1}^{m} \sim p(\#z)$ a batch of latent variables; $\varepsilon \sim U[0, 1]$.
                \STATE $\widetilde{\#x} \leftarrow G_{\theta}(\#z)$
                \STATE $\hat{\#x} \leftarrow \varepsilon\#x + (1-\varepsilon)\widetilde{\#x}$
                \STATE $L^{(i)} \leftarrow D_{w}(\widetilde{\#x}) - D_{w}(\#x) + \lambda (\| \nabla_{\hat{\#x}}D_{w}(\hat{\#x}) \|_2  - 1)^2$
    	    \ENDFOR
    	    \STATE $w \leftarrow \mathrm{Adam}(\nabla_{\#x} \frac{1}{m} \sum_{i=1}^{m} L^{(i)}, w, \alpha, \beta_1, \beta_2$)
        \ENDFOR
        \STATE Sample $\{ \#z^{(i)} \}_{i=1}^{m} \sim p(z)$ a batch of latent variables.
        \STATE $\theta \leftarrow \mathrm{Adam}(\nabla_{\theta} \frac{1}{m} \sum_{i=1}^{m} -D_{w}(G_\theta(\#z)), \theta, \alpha, \beta_1, \beta_2$)
    \ENDWHILE
    \end{algorithmic}
\end{algorithm}

\paragraph*{Boundary-Seeking Generative Adversarial Networks (BGAN)} were introduced in~\cite{BSGAN}. Similarly as WGAN, BGAN are focusing on GAN learning, they train a generator to match the target distribution that converges to the data distribution at the limit of a perfect discriminator. The interpretation of this is training the generator in order to produce the samples lying on the decision boundary of the current discriminator (which explains the name ``Boundary-Seeking'' GAN).

We omit the details of BGAN here and refer the reader to the original BGAN paper~\cite{BSGAN}. The authors claim that it produces similar results as WGAN, which is proved to be more stable against the mode collapse. We list BGAN here because of its main strength, which is a definition of a unified learning framework for both discrete and continuous variables, and interesting intuition about the samples lying on the decision boundary of the discriminator.

\subsection{GAN: Recent Advances}


In this section, some of the related and recent GAN advances are briefly described. BigGAN \cite{brock2018large}, a class-conditional image synthesis approach, achieved a new level of performance for the large scale of ImageNet GAN models while allowing fine control over the trade-off between sample fidelity and variety by truncating the latent space. Moreover the authors presented an analysis of the training stability.
A new training methodology for GAN is proposed in \cite{karras2017progressive} where the generator and discriminator are made larger starting from a low resolution and adding new layers progressively, leading to generating a high resolution, high quality and fine details outputs. The authors described different implementation details to discourage the competition between the generator and discriminator. Furthermore, the proposed progressive training is showed to speed up and stabilize the training process. 
A novel generator architecture for GAN based on style transfer approaches is proposed in \cite{karras2018style}, leading to unsupervised separation of attributes and stochastic effects, that enabled fine control based on different levels of the generated images. Moreover, the experiments showed better interpolation properties. 
Self-Attention Generative Adversarial Network (SAGAN) \cite{zhang2018self} integrates a self-attention mechanism into GAN framework that allows attention-driven modeling for image generation where details are generated using cues from all feature locations. Moreover, it was shown that spectral normalization applied to
the generator stabilizes the GAN training process.

\begin{table} 
	\centering 
    \caption{Summary of the prominent GAN derivatives.} \label{tab:gan_derivatives}
    \noindent
    \begin{tabular}{|p{0.30\linewidth}|p{0.60\linewidth}|}
         \hline
         \textbf{GAN Derivative} & \textbf{References} \\
         \hline
         \hline
         General functionality & Vanilla GAN~\cite{GAN}, Conditional GAN (CGAN)~\cite{CGAN} \\ 
         \hline 
         Stabilizing functionality & Wasserstein GAN (WGAN)~\cite{WassersteinGAN}, Improved WGAN (iWGAN)~\cite{Gulrajani-NIPS-2017}, Boundary-Seeking GAN (BGAN)~\cite{BSGAN} \\
         \hline
    \end{tabular}
\end{table}

%% file: include/ganapplicationsforad.tex
\section{GAN APPLICATIONS FOR AUTONOMOUS DRIVING} \label{sec:gan4ad}

In this section, we discuss the potential domains of application of GAN (in the, basically, arbitrary variant) with a focus on autonomous driving. Thanks to the increasing popularity of GAN, a lot of applications have been already identified. Most of them are, based on the GAN nature, related to the image-to-image translation or possibilities of the semi/unsupervised learning. We further discuss the following main categories: 1) advanced data augmentation (which is split into several subcategories, such as $2$D \& $3$D synthesis, video synthesis, super resolution, or inpainting); 2) semi-supervised/unsupervised learning; 3) loss function learning; 4) adversarial training \& testing. A short description of each of these categories follows. For the reader's convenience, we also summarized all key references in a Table~\ref{tab:gan_applications}.



\subsection{Advanced Data Augmentation}

%

Data Augmentation is a natural application of GAN. There are numerous papers dedicated to Image-to-Image translation on the top-level conferences from last few years (\cite{Isola-CVPR-2017,Zhu-ICCV-2017,Liu-NIPS-2017}, to name a few). GAN create realistic looking images, automatic conversion from a black and white image to a colored one, areal image to map, edges to a photo-realistic images of the sketched objects, or even some advanced stuff like day to night or summer to winter, or context-aware object placement~\cite{Lee-NIPS-2018} (which is a very interesting application for autonomous driving, giving us the possibility to enhance existing dataset by realistic looking images which we are lacking--- e.g. small number of images with pedestrians, etc.). A natural extension of the Image-to-Image translation is a Video-to-Video translation, where the idea remains the same. However, the task is much more difficult thanks to the temporal information, which also has to remain consistent. 

In~\cite{Isola-CVPR-2017}, the authors formulate the problem of Image-to-Image translation as an instance of CGAN, where the objective loss, inspired by~\cite{Pathak-CVPR-2016}, is mixed with a traditional L$1$ loss. Note, that within such formulation the discriminator's task remains unchanged. However, the generator's task, besides fooling the discriminator, is to produce samples near the ground-truth in the L$1$ sense. The authors argue, that using L$1$ leads to less blurring, compared to the L$2$. The $\mathcal{L}_{L1}$ term is formulated as follows: 
\begin{equation}
	\mathcal{L}_{L1} = \mathbb{E}_{\#x, \#y \sim p_{\mathrm{data}}(\#x, \#y), \#z \sim p_{\#z}(\#z)} \big[ \left\| \#y - G(\#x, \#y) \right\|_1  \big] \;.
\end{equation}
If we denote the equation~(\ref{eq:CGAN}) as $\mathcal{L}_{\mathrm{CGAN}}(G, D)$, the final objective of~\cite{Isola-CVPR-2017} can be expressed as:
\begin{equation}
	G^* = \arg\min_{G}\max_{D} \mathcal{L}_{\mathrm{CGAN}}(G, D) + \lambda \mathcal{L}_{L1}(G) \;,
\end{equation}
where $\lambda$ tells how much we want the $\mathcal{L}_{L1}$ loss influence the objective. Even though the results look impressive at the first sight, we should point out here, that there are certain artifacts appearing basically in all of the examples. 

The authors of~\cite{Isola-CVPR-2017} continued working on the Image-to-Image translation topic, and in a follow-up paper~\cite{Zhu-ICCV-2017}, they move further, by introducing an algorithm for an unpaired Image-to-Image translation. Briefly, the authors are learning a mapping $G \colon X \rightarrow Y$, such that the distribution of images from $G(X)$ is indistinguishable from the distribution $Y$. Because such mapping is highly under-constrained, they couple it with an inverse mapping $F \colon Y \rightarrow X$ and introduce a cycle consistent loss enforcing $F(G(X)) \approx X$, and vice versa. The proposed algorithm is called CycleGAN. 

Because of the inverse mapping, the CycleGAN uses a slightly different definition of $\mathcal{L}_{\mathrm{GAN}}$ than what we introduced here:
\begin{eqnarray}
	\mathcal{L}_{\mathrm{GAN}}(G, D_Y, X, Y) \!\!\!\! &=& \!\!\!\! \mathbb{E}_{\#y \sim p_{\mathrm{data}}(\#y)} \big[\log D_Y(\#y)\big] \nonumber \\
    \!\!\! & & \!\!\! + \mathbb{E}_{\#x \sim p_{\mathrm{data}}(\#x)} \big[ \log (1-D_Y(G(\#x)) \big] \;, \nonumber \\
\end{eqnarray}
The cycle consistency loss is defined as follows:
\begin{eqnarray}
	\mathcal{L}_{\mathrm{cyc}}(G, F) & = & \mathbb{E}_{\#x \sim p_{\mathrm{data}}(\#x)} \big[ \|F(G(\#x)) - \#x \|_1 \big] \nonumber \\
	& & + \mathbb{E}_{\#y \sim p_{\mathrm{data}}(\#y)} \big[ \|G(F(\#y)) - \#y \|_1 \big] \;.
\end{eqnarray}
The full objective of CycleGAN is defined as 
\begin{eqnarray}
	\mathcal{L}(G, F, D_X, D_Y) &=& \mathcal{L}_{\mathrm{GAN}}(G, D_Y, X, Y) \nonumber \\ 
    & & + \mathcal{L}_{\mathrm{GAN}}(F, D_X, Y, X) \nonumber \\
    & & + \lambda \mathcal{L}_{\mathrm{cyc}}(G, F) \;,
\end{eqnarray}
where again $\lambda$ controls the relative importance of the objective components. The full optimization task is simply
\begin{equation}
	G^*, F^* = \arg\min_{G, F}\max_{D_X, D_Y} \mathcal{L}(G, F, D_X, D_Y) \;.
\end{equation}
As the main benefit compared to the previous work~\cite{Isola-CVPR-2017}, the authors consider the ability of CycleGAN to operate without a specific supervision. However, the main drawback remains unsolved--- the generated images, after a careful inspection, show the same artifacts as the previous work. On the other hand, CycleGAN showed even more advanced examples of possible data augmentation (changing of an object class, like apples $\leftrightarrow$ oranges, or zebras $\leftrightarrow$ horses).


Since both~\cite{Isola-CVPR-2017}, and~\cite{Zhu-ICCV-2017} are based on GAN, they also require some compromises in the network architecture for both $D$, and $G$. It might be interesting to try to adopt the ideas and using WGAN instead. 

Slightly different approach than CycleGAN is described in~\cite{Liu-NIPS-2017}. They also deal with an UNsupervised Image-to-image Translation (UNIT). However, instead of explicitly modeling the cycle consistency, they use an assumption of shared-latent space, which, as the authors show, implies the cycle consistency constraint. The proposed UNIT framework is based on GAN and variational autoencoders (VAE). We omit the technical details about UNIT here and refer the reader to the original paper~\cite{Liu-NIPS-2017}. The results are similar to those presented by CycleGAN, including the artifacts appearing in the generated images. 


In \cite{Odena-ICML-2017}, an algorithm for conditional image synthesis with auxiliary classifier GAN (AC-GAN) is introduced. The authors claim that even though structurally AC-GAN are not much different from the existing models, they seem to stabilize the training. Despite the AC-GAN formulation, the paper focuses on measuring the extent to which a model make use of its given output resolution, and on measuring the perceptual variability of samples generated by the model. In these areas, this work can be proclaimed as a pioneer. \\




\noindent \textbf{Synthesis:}
Data Synthesis using GAN can be exploited for different autonomous driving applications. For example, data augmentation to enable better generalization over different weather and lighting conditions conducting visual perception by generating abstracted views such as semantic segmentation of input camera frames. In addition to sensor correcting such as fixing noisy inputs and sensor modeling. Synthesis can be achieved spatially in two and three dimensional spaces, in addition to spatio-temporal spaces such as videos.  \\

\begin{figure*}
    \centering
    \includegraphics[width=\linewidth]{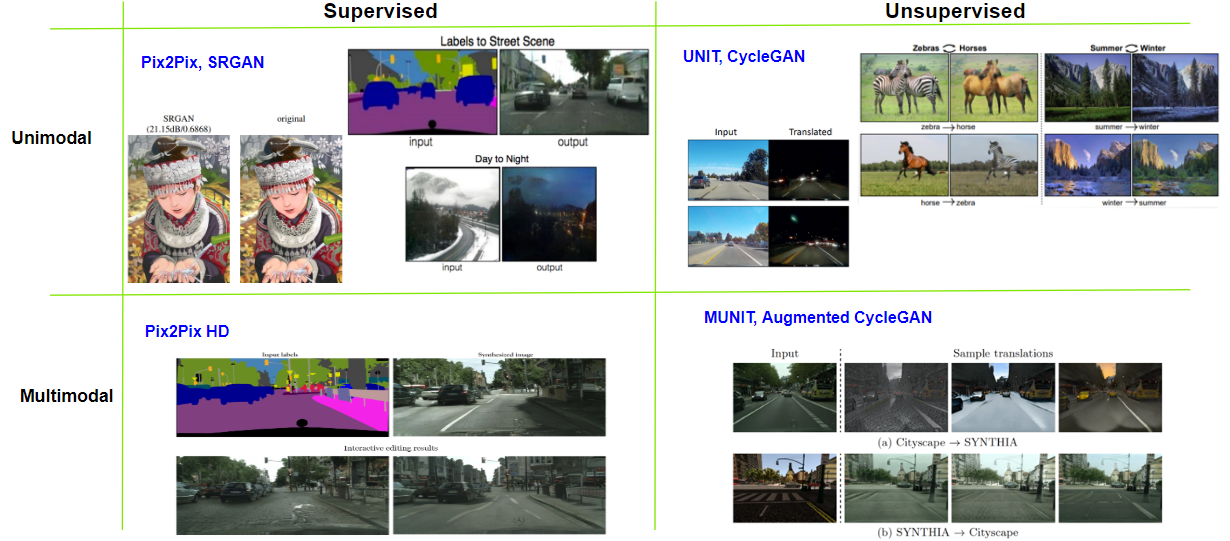}
    \caption{Successful application of GAN for autonomous driving - Image to Image Translation}
    \label{fig:image_translation}
\end{figure*}

\begin{table} 
	\centering 
    \caption{Summary of applications of GAN for autonomous driving.} \label{tab:gan_applications}
    \noindent
    \begin{tabular}{|p{0.30\linewidth}|p{0.60\linewidth}|}
         \hline
         \textbf{GAN Applications} & \textbf{References} \\
         \hline
         \hline
         2D Synthesis & Pix2Pix \cite{Isola-CVPR-2017}, SRGAN \cite{ledig2017photo},  CycleGAN \cite{Zhu-ICCV-2017}, DiscoGAN \cite{kim2017learning}, StarGAN \cite{choi2018stargan}, UNIT \cite{liu2017unsupervised}, Pix2PixHD \cite{wang2018high}, BicycleGAN \cite{zhu2017toward}, MUNIT \cite{huang2018multimodal}, Augmented GAN \cite{almahairi2018augmented}\\
         \hline
         3D Synthesis & 3D-GAN\cite{wu2016learning}, PrGAN \cite{gadelha20173d}, PC-GAN \cite{li2018point} \\
         \hline
         Video Synthesis & video-to-video \cite{wang2018video}, TGAN \cite{saito2017temporal}, \cite{vondrick2016generating}\\
         \hline
         Domain Adaptation & Pixel level \cite{bousmalis2017unsupervised}, GraspGAN \cite{bousmalis2017using}, \cite{pan2017virtual}, \cite{bewley2018learning}\\
         \hline
         Object Detection & SeGAN \cite{ehsani2018segan}, \cite{wang2017fast}, Perceptual-GAN \cite{li2017perceptual}\\
         \hline
         Super Resolution & SRGAN \cite{ledig2017photo}\\
         \hline
         Inpainting & \cite{iizuka2017globally}, \cite{yeh2017semantic}, \cite{pathak2016context} \\
         \hline 
         Advanced Data Augmentation & Context-aware Synthesis and Placement of Object Instances~\cite{Lee-NIPS-2018} \\
         \hline
    \end{tabular}
\end{table}

\noindent \textbf{2D Synthesis:}
GAN have been proposed as a generative framework that maps random noise to synthetic, realistically looking images following the training data distribution. In image-to-image translation, conditional GAN are adopted to enable the generative framework to condition both the generator and the discriminator on prior knowledge. In this case, the model is trained to map from images in a source domain to images in a target domain. Image-to-Image translation can be approached based on two main directions: 1) paired or unpaired; 2) unimodal or mutilmodal. 
In the unimodal paired image translation, for example Pix2Pix~\cite{Isola-CVPR-2017}, SRGAN~\cite{ledig2017photo}, the model learns to map images where the training data is organized in pairs of input and output samples. In many cases, the paired training data could not be available. In the unimodal unpaired approach, for example CycleGAN~\cite{Zhu-ICCV-2017}, DiscoGAN~\cite{kim2017learning}, StarGAN~\cite{choi2018stargan}, UNIT~\cite{liu2017unsupervised}, the image translation is conducted on unpaired data from two domains, where it learns a mapping between the two domains without supervision. In multimodal image translation, it is possible to generate several images of different styles based on a single source image. Multimodal translation can be paired Pix2PixHD~\cite{wang2018high}, BicycleGAN~\cite{zhu2017toward}, or unpaired MUNIT~\cite{huang2018multimodal}, Augmented GAN~\cite{almahairi2018augmented}. \\ 

\noindent \textbf{3D Synthesis:}
Recently, one of the essential sensors used for automated driving is LiDAR, mainly because of its physical ability to perceive accurate depth and to produce $3$D point clouds, regardless of lighting conditions. Most of GAN approaches are not applicable to $3$D point clouds, however Point Cloud GAN (PC-GAN)~\cite{li2018point} proposed a two fold modification to GAN algorithm for learning to generate point clouds. Moreover, studies were provided for transforming images into point clouds. 3D-GAN framework is introduced in~\cite{wu2016learning} to map from a low-dimensional probabilistic space to the space of $3$D objects. PrGAN~\cite{gadelha20173d}, investigated the task of generating a distribution over $3$D structures given $2$D views of multiple objects taken from unknown viewpoints. This approach produces $3$D shapes of comparable quality to GAN trained on $3$D data, allows generating new views from an input image in an unsupervised manner. \\
    
\noindent \textbf{Video Synthesis:}
The goal of video-to-video synthesis is to learn a mapping from an input video to a realistic output video. Simply applying image to image synthesis results in temporally incoherent videos. Recently, a video synthesis approach based on GAN framework is proposed in~\cite{wang2018video} where a spatio-temporal adversarial objective is used to synthesis $2$k resolution videos of street scenes up to $30$ seconds long. This allows developers and artists to create new interactive $3$D virtual worlds for different domains including automotive domain. 
Temporal GAN (TGAN)~\cite{saito2017temporal}, on the other hand, learns a semantic representation of unlabeled videos and generates videos, using a temporal generator and an image generator.   
Earlier, a GAN network for video with a spatio-temporal convolutional architecture is proposed in~\cite{vondrick2016generating} where the foreground of the scene is separated from the background, generating small one second videos. \\

\noindent \textbf{Domain adaptation from simulation to real:}
Autonomous driving systems usually require collecting and annotating a lot of training data. On the other hand, using simulated environments enables much easier collection, but models trained on simulated environments often fail to generalize on real environments. Domain adaptation allows a machine learning model trained on samples from a source domain to generalize to a target domain. 
GAN based Pixel-Level domain adaptation method is proposed in~\cite{bousmalis2017unsupervised} where the adaptation process showed to produce plausible samples and to generalize well to object classes unseen during the training. GraspGAN~\cite{bousmalis2017using} extended the pixel-level domain adaptation to reduce the number of real world samples needed by up to $50$ times for vision-based grasping system.
Reinforcement Learning for Autonomous Driving model trained in virtual environment is shown to perform well in real environment~\cite{pan2017virtual}. Two image-to-image translation networks are used. The first network translates virtual images to their segmentation, the second network translates segmented images into their realistic counterpart. Accordingly, the driving policy can be easily adapted to real environment.
A method for transferring a vision-based
lane following driving policy from a simulated to a real environment is presented in~\cite{bewley2018learning} where a model for end-to-end driving is constructed by learning to translate between simulated and real images, jointly learning a control policy from this common latent space using labels from an expert driver in the simulated environment. It was shown that the proposed system is capable of leveraging simulation to learn a driving policy to directly transfer to real world scenarios. \\

\noindent \textbf{Object Detection:}
In real life situations, especially for autonomous driving, objects often occlude each other and inferring the occluded  objects is essential for scene understanding and taking decisions. SeGAN~\cite{ehsani2018segan} is an approach for both segmentation and generation of the occluded parts of objects, where the proposed network has three parts: segmentor, generator, and discriminator. 
On the other hand, some occlusions are very rare in the training data. Learning a model invariant to such occurrences is proposed in~\cite{wang2017fast} where an adversarial network learns to generate challenging occlusions and deformations examples.
Detection of small objects is often challenging task. Perceptual-GAN~\cite{li2017perceptual} model narrows representation difference of small and large objects, where the generator learns to transfer the small objects representations large ones to fool the competing discriminator. \\

\noindent \textbf{Super Resolution:}
In autonomous driving domain, low resolution sensors may be used. Generating the corresponding high resolution representations could enable and enhance the systems that were trained on high resolution inputs. However, estimating a high resolution representation from a low resolution counterpart is a highly challenging task. 
SRGAN \cite{ledig2017photo} is a generative adversarial network for image super-resolution framework able to infer photo-realistic natural images for 4x upscaling factors. A perceptual loss function is proposed consisting of both an adversarial loss for natural output, and a content loss for perceptual similarity.  \\

\noindent \textbf{Inpainting:}
In real life autonomous driving situations, sensors may read noisy data or may suffer from failures causing incomplete readings. Inpainting can provide a solution. Globally and locally consistent image completion approach is proposed in \cite{iizuka2017globally} where global and local context discriminators are trained to distinguish real images from completed ones. The global discriminator is used for the entire coherent of the generated image and the local discriminator ensures the local consistency. The approach showed to naturally complete the missing fragments. Inpainting is a challenging task especially for large missing parts. \cite{yeh2017semantic} proposed a method for semantic image inpainting by conditioning on the available data making the inference possible irrespective of how the missing parts are structured.  
\cite{pathak2016context} introduced context encoders that predict missing parts of a scene from their surroundings. Using an adversarial loss is found to produce sharper results. Moreover, the network learns a representation that captures both the appearance and the semantics of the scene. 

\subsection{Semi-supervised/Unsupervised Learning}

Unsupervised learning is the holy grail of machine learning. It is the hardest scenario of machine learning, as it requires just a bunch of unannotated data, and the output is a learned meaningful model, which pursues a certain task (which can be inferred from the data). 

Recently, one of the most popular approaches to unsupervised learning are Variational Autoencoders (VAE). They are built on top of the standard function approximators (neural networks) and can be trained by stochastic gradient descent. Our description of VAE framework follows~\cite{Doersch2016TutorialOV}. 

In principle, VAE is aiming to maximize the probability of each $\#x$ in the training set under the generative process, according to
\begin{equation}
	p(\#x) = \int p(\#x | \#z;\ \theta) p(\#z) d\#z \; \label{eq:vae_px}
\end{equation}
where $\#z \in \mathcal{Z}$ is a \textit{latent} variable specifying the class of the generated object.
The intuition behind this framework is nothing else than maximum likelihood principle--- if the model is likely to produce training set samples, then it is also likely to produce similar samples, and unlikely to produce dissimilar ones. VAE approximately maximizes the probability function. The name ``autoencoders'' comes from the training objective, which, derived from this setup, have an encoder and a decoder, and resembles a traditional autoencoder. The benefit of VAE is that we can sample from $p(\#x)$ (without performing Markov Chain Monte Carlo). 

To solve the maximization, VAE have to deal with the problem of how to define latent variables $\#z$ (i.e. decide what information they represent), as well as how to deal with the integral over $\#z$. 

VAE assume that there is no simple interpretation of dimensions of $\#z$, rather they claim that samples of $\#z$ might be drawn from a simple distribution, i.e. $\mathcal{N}(0; \mathcal{I})$, where $\mathcal{I}$ is the identity matrix (notice, that any distribution in $d$ dimensions can be generated by taking a set of d variables that are normally distributed, and map them through a sufficiently complicated function). Conceptually, the approximation of $p(\#x)$ is straightforward— one just needs to sample a large number of samples $\{\#z_1, \dots, \#z_n \}$, and then compute $p(\#x) \approx \frac{1}{n} \sum_{i=1}^{n} p(\#x|\#z_i)$. A problem arises in high dimensional spaces because then n might be extremely large, to get an accurate estimate of $p(\#x)$. 

The key idea behind VAE is to attempt to sample values of $\#z$ that are likely to produce $X$ and compute $P(X)$ just from those. For this purpose, let us define a new function $Q(\#z|X)$, which takes a value of $X$ and give us a distribution over $\#z$ values, that are likely to produce $X$. Hopefully, the space of $\#z$ values that are likely under $Q$ will be much smaller than the space of all $\#z$'s, that are likely under the prior $P(\#z)$. This lets us compute $\mathbb{E}_{\#z \sim Q} P(X|\#z)$. However, if $\#z$ is sampled from an arbitrary distribution with a probability density function $Q(\#z)$ (i.e. which is not necessarily $\mathcal{N}(0, I)$), we need to relate $\mathbb{E}_{\#z \sim Q} P(X|\#z)$ with $P(X)$. By defining a Kullback-Leibler divergence between $P(\#z|X)$ and $Q(\#z)$, applying a Bayes rule to $P(\#z|X)$, and some basic algebra, we finally arrive at a core equation of VAE:
\begin{eqnarray}
	\log P(X) - \mathrm{KL}[Q(\#z|X)\|P(\#z|X)] =  \mathbb{E}_{\#z \sim Q} \left[ \log P(X|\#z) \right] \nonumber \\
    -\mathrm{KL}[Q(\#z|X)\|P(\#z)] \;. \label{eq:VAE_autoencoder}
\end{eqnarray}
The left hand side is the quantity we want to maximize ($\log P(X)$ plus an error term making $Q$ produce $\#z$'s that can reproduce a given $X$; this term should be small if $Q$ is high-capacity). The right hand side is optimizable via stochastic gradient descent, given the right choice of $Q$. Note, that the right hand side already looks like an autoencoder, since $Q$ is ``encoding'' $X$ into $\#z$ and $P$ is ``decoding'' it to reconstruct $X$.

In~(\ref{eq:VAE_autoencoder}), we are maximizing $\log P(X)$, while simultaneously minimizing $\mathrm{KL}[Q(\#z|X)\|P(\#z|X)]$. The second term on the left-hand side is pulling $Q(\#z|X)$ to match $P(\#z|X)$. With an assumption of an arbitrarily high-capacity model for $Q(\#z|X)$ it will actually match $P(\#z|X)$, in which case the KL divergence term will be zero, and we will be optimizing $\log P(X)$ directly.

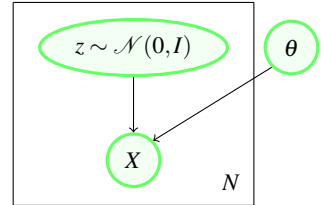
\begin{figure}[tbp]
	\centering
    \begin{tikzpicture}[
    roundnode/.style={circle, draw=green!60, fill=green!5, very thick, minimum size=7mm},
    squarednode/.style={rectangle, draw=red!60, fill=red!5, very thick, minimum size=5mm},
    ellipsenode/.style={ellipse, draw=green!60, fill=green!5, very thick, minimum size=7mm},
    ]
        \node[roundnode] (x) at (0, 0) {$X$};
        \node[ellipsenode] (z) at (0, 1.5) {$\#z \sim \mathcal{N}(0, I)$};
        \node[roundnode] (theta) at (2.1, 1.5) {$\theta$};

        \draw[-] (-1.6, 2.1) -- (-1.6, -0.6);
        \draw[-] (-1.6, -0.6) -- (1.6, -0.6);
        \draw[-] (-1.6, 2.1) -- (1.6, 2.1);
        \draw[-] (1.6, 2.1) -- (1.6, -0.6);

        \node[] (N) at (1.3, -0.3) {$N$};

        \draw[->] (z.south) -- (x.north);
        \draw[->] (theta.south west) -- (x.north east);

    \end{tikzpicture}
	\caption{Standard VAE model represented as a graphical model. The rectangle is a ``plate notation'', meaning that we can sample from $\#z$ and $X$ $N$ times, while the model parameters $\theta$ remain fixed.} \label{fig:VAE_graphical_model}
\end{figure}

\begin{figure*}
	\centering
    \begin{tikzpicture}[
    	roundnode/.style={circle, draw=black!60, fill=black!5, very thick, minimum size=7mm},
    	redsquarednode/.style={rectangle, draw=red!60, fill=red!5, very thick, minimum size=5mm, text=red},
        squarednode/.style={rectangle, draw=black!60, fill=black!5, very thick, minimum size=5mm},
        bluesquarednode/.style={rectangle, draw=blue!60, fill=blue!5, very thick, minimum size=5mm, text=blue},
    	ellipsenode/.style={ellipse, draw=green!60, fill=green!5, very thick, minimum size=7mm},
    ]
    	\node[squarednode] (X) at (0, 0) {$X$};
        \node[squarednode] (encoder) at (0, 1) {Encoder ($Q$)};
        \node[squarednode] (mux) at (-0.6, 2) {$\mu (X)$};
        \node[squarednode] (sigmax) at (0.6, 2) {$\Sigma (X)$};
        \node[redsquarednode] (sample) at (3, 3) {Sample $\#z$ from $\mathcal{N}(\mu (X), \Sigma (X))$};
        \node[bluesquarednode] (KL) at (-1, 4) {$\mathrm{KL}[\mathcal{N}(\mu(X),\Sigma(X))\|\mathcal{N}(0, I)]$};
        \node[squarednode] (decoder) at (3, 5) {Decoder ($P$)};
        \node[squarednode] (fz) at (3, 6) {$f(\#z)$};
        \node[bluesquarednode] (xdifff) at (3, 7) {$\| X - f(\#z) \|^2$};
        
        \draw[->] (X.north) -- (encoder.south);
        \draw[->] (encoder.north) -- (mux.south);
        \draw[->] (encoder.north) -- (sigmax.south);
        \draw[->] (sigmax.north) -- (sample.south);
        \draw[->] (sigmax.north) -- ([xshift=3em]KL.south);
        \draw[->] (mux.north) -- ([xshift=-4em]sample.south);
        \draw[->] (mux.north) -- (KL.south);
        \draw[->] (sample.north) -- (decoder.south);
        \draw[->] (decoder.north) -- (fz.south);
        \draw[->] (fz.north) -- (xdifff.south);
        
        \draw[-,dashed,gray] (5.75,-0.3) -- (5.75,7.5);
        
        \node[squarednode] (X2) at (8, 0) {$X$};
        \node[squarednode] (encoder2) at (8, 1) {Encoder ($Q$)};
        \node[squarednode] (mux2) at (7.4, 2) {$\mu (X)$};
        \node[squarednode] (sigmax2) at (8.6, 2) {$\Sigma (X)$};
        \node[roundnode] (asterisk) at (12,2) {*};
        \node[roundnode] (plus) at (12,3.5) {+};
        \node[redsquarednode] (sample2) at (12, 1) {Sample $\epsilon$ from $\mathcal{N}(0, I)$};
        \node[bluesquarednode] (KL2) at (8.5, 5) {$\mathrm{KL}[\mathcal{N}(\mu(X),\Sigma(X))\|\mathcal{N}(0, I)]$};
        \node[squarednode] (decoder2) at (12, 5) {Decoder ($P$)};
        \node[squarednode] (fz2) at (12, 6) {$f(\#z)$};
        \node[bluesquarednode] (xdifff2) at (12, 7) {$\| X - f(\#z) \|^2$};
        
        \draw[->] (X2.north) -- (encoder2.south);
        \draw[->] (encoder2.north) -- (mux2.south);
        \draw[->] (encoder2.north) -- (sigmax2.south);
        \draw[->] (sigmax2.east) -- (asterisk.west);
        \draw[->] (sigmax2.north) -- ([xshift=0.25em]KL2.south);
        \draw[->] (mux2.north) -- (plus.west);
        \draw[->] (mux2.north) -- ([xshift=-3.25em]KL2.south);
        \draw[->] (sample2.north) -- (asterisk.south);
        \draw[->] (decoder2.north) -- (fz2.south);
        \draw[->] (fz2.north) -- (xdifff2.south);
        \draw[->] (asterisk.north) -- (plus.south);
        \draw[->] (plus.north) -- (decoder2.south);
        
    \end{tikzpicture}
    \caption{A training-time variational autoencoder implemented as a feed-forward neural network, where $P(X|\#z)$ is Gaussian. Red shows non-differentiable sampling operations. Blue shows loss layers. The left diagram is without using a reparametrization trick, while the right is with it. The feed-forward behavior of both networks is identical. However, the backpropagation algorithm can be applied only to the one on the right. Taken from~\cite{Doersch2016TutorialOV}.} \label{fig:VAE_traiing_scheme}
\end{figure*}
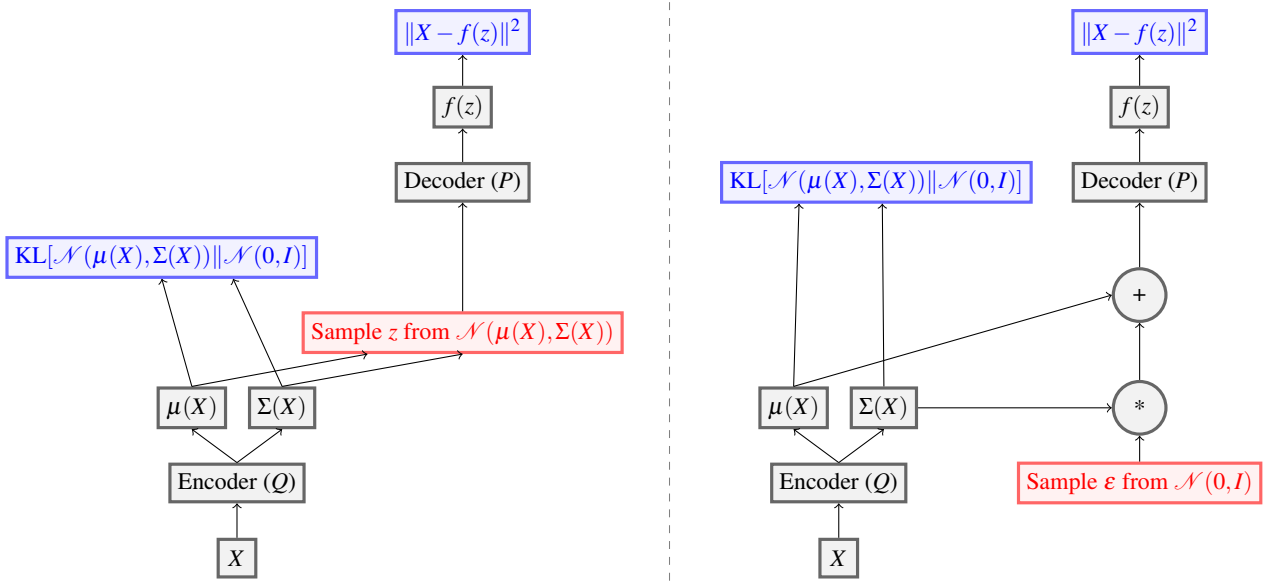

\begin{figure}
	\centering
    \begin{tikzpicture}[
    	roundnode/.style={circle, draw=black!60, fill=black!5, very thick, minimum size=7mm},
    	redsquarednode/.style={rectangle, draw=red!60, fill=red!5, very thick, minimum size=5mm, text=red},
        squarednode/.style={rectangle, draw=black!60, fill=black!5, very thick, minimum size=5mm},
        bluesquarednode/.style={rectangle, draw=blue!60, fill=blue!5, very thick, minimum size=5mm, text=blue},
    	ellipsenode/.style={ellipse, draw=green!60, fill=green!5, very thick, minimum size=7mm},
    ]
    	\node[redsquarednode] (sample) at (0, 0) {Sample $\#z$ from $\mathcal{N}(0, I)$};
        \node[squarednode] (decoder) at (0, 1) {Decoder ($P$)};
        \node[squarednode] (fz) at (0, 2) {$f(\#z)$};
        
        \draw[->] (sample.north) -- (decoder.south);
        \draw[->] (decoder.north) -- (fz.south);
        
    \end{tikzpicture}
    \caption{The testing-time variational autoencoder allowing us to generate new samples. The ``encoder'' pathway is simply discarded.} \label{fig:VAE_test}
\end{figure}
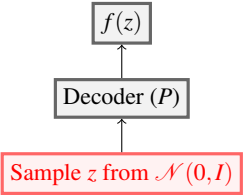

In order to make VAE work, it is essential to drive $Q$ to produce codes for $X$, that $P$ can reliably decode. See Figure~\ref{fig:VAE_traiing_scheme} left, for a different view on the problem. The forward pass of this network works fine. However, the backpropagation of the error through a layer sampling $\#z$ from $Q(\#z|X)$ is not possible, since this is a non-continuous operation and therefore has no gradient. The solution is a so-called ``reparametrization trick''~\cite{Kingma-ICLR-2014}, which moves the sampling to an input layer. Figure~\ref{fig:VAE_traiing_scheme} right depicts the reparametrization scheme. We should point out that this trick works only if sampling from $Q(\#z|X)$ is possible through evaluation of a continuous function $h(\eta, X)$ in X, where $\eta$ is a noise from a distribution that is not learned. This basically means that $Q(\#z|X)$ (and therefore also $P(\#z)$) cannot be a discrete distribution.
At the test time, when we want to generate new samples, we simply input values of $\#z \sim \mathcal{N}(0, I)$ into the decoder. This is schematically shown in Figure~\ref{fig:VAE_test}.

We will skip the definition of Conditional VAE here since the extension is quite straightforward, and refer the reader to the VAE tutorial paper~\cite{Doersch2016TutorialOV} for details.

In~\cite{Rosca-ArXiv-2017}, the authors show how to combine GAN and VAE in such a way, that the best of both worlds is used, and the limitations of both methods are mitigated. They propose to use a hybrid loss function which combines VAE and GAN:
\begin{eqnarray}
	\mathcal{L}(\theta, \eta) & = & \mathbb{E}_{q_\eta(\#z|\#x)} \Bigg[ -\lambda \| \#x - \mathcal{G}_{\theta}(\#z) \|_1 \nonumber \\
   & & + \log \frac{\mathcal{D}_\phi(\mathcal{G}_{\theta}(\#z))}{1 - \mathcal{G}_{\theta}(\#z))}
   + \log\frac{\mathcal{C}_{\#\omega}(\#z)}{1 - \mathcal{C}_{\#\omega}(\#z)} \Bigg] \;. \label{eq:alphaGAN}
\end{eqnarray}
We need to build four networks here. The classifier $\mathcal{D}_\theta(\#x)$, which is trained to discriminate between reconstructions from an auto-encoder, and real data points. A second classifier, which is trained to discriminate between latent samples produced by the encoder and samples from a standard Gaussian. The deep generative model $\mathcal{G}_\theta(\#z)$. And also the encoder network $q_\eta(\#z|\#x)$, which can be implemented using arbitrary deep network. The authors refer to~(\ref{eq:alphaGAN}) as $\alpha$-GAN. The algorithm alternates between updates of parameters of the generator $\theta$, encoder $\eta$, synthetic likelihood discriminator $\phi$, and the latent code discriminator $\#\omega$. For the detailed description of the algorithm, we refer the reader to the original paper~\cite{Rosca-ArXiv-2017}.

The problem of semi-supervised learning is covered in~\cite{Kingma-NIPS-2014}, which shows, that deep generative models and approximate Bayesian inference exploiting recent advances in variational methods can be used to provide significant improvements, making generative approaches highly competitive for semi-supervised learning. The authors describe a new framework for semi-supervised learning with generative models, employing rich parametric density estimators formed by the fusion of the probabilistic modeling and deep neural networks.

Interesting view of unsupervised learning of visual representations is shown in~\cite{Noroozi-ECCV-2016}, where the problem is posed as solving a jigsaw puzzle. The authors introduce the context-free network (CFN), which takes image tiles as input and explicitly limits the receptive field. The experimental evaluation shows that the learned features capture semantically relevant content. 

Authors of~\cite{Erhan-JMLR-2010} argue that unsupervised pre-training is beneficial for deep learning in general. Their results suggest, that unsupervised pre-training guides the learning 
6
towards basins of attraction of minima that support a better generalization from the training dataset.

\subsection{Learned Loss Functions}

Last but not least domain of GAN applications is its implicit ability of learning non-trivial loss functions. In certain setup, the GAN optimization criterion expresses the loss function indirectly, and therefore by optimizing the complex GAN criterion, we are also optimizing a loss function which we do not have to formulate explicitly. Unlike the previous domains, this is so far the least explored topic, so there are not many papers available in this area. 

In~\cite{Santos-2017}, the authors propose discriminative adversarial networks (DAN) for semi-supervised learning and loss function learning\footnote{We should point out that in~\cite{Santos-2017} the DAN is used on two different tasks of Natural Language Processing field. However, there do not seem to be any burdens prohibiting usage in the computer vision field.}. Unlike the vanilla GAN, DAN uses two discriminators, instead of a generator and a discriminator. It can be seen as a framework to learn a loss function for predictors, that also implements semi-supervised learning. 

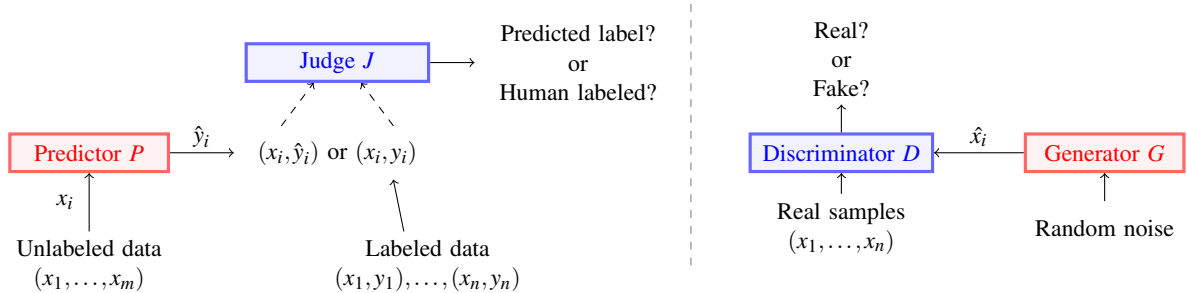
\begin{figure*}
	\centering
    \begin{tikzpicture}[
    	roundnode/.style={circle, draw=black!60, fill=black!5, very thick, minimum size=7mm},
    	redsquarednode/.style={rectangle, draw=red!60, fill=red!5, very thick, minimum size=5mm, text=red, text width=6em, text centered},
        squarednode/.style={rectangle, draw=black!60, fill=black!5, very thick, minimum size=5mm},
        bluesquarednode/.style={rectangle, draw=blue!60, fill=blue!5, very thick, minimum size=5mm, text=blue, text width=7em, text centered},
    	ellipsenode/.style={ellipse, draw=green!60, fill=green!5, very thick, minimum size=7mm},
        emptynode/.style={draw=none, fill=none, minimum size=7mm, text centered, text width=8em},
    ]
    	\node[redsquarednode] (predictor) at (0,0) {Predictor $P$};
        \node[bluesquarednode] (judge) at (3.3, 1.2) {Judge $J$};
        \node[emptynode] (xypairs) at (3.3, 0) {$(\#x_i, \hat{\#y}_i)$ or $(\#x_i, \#y_i)$};
        \node[emptynode] (unlabeleddata) at (0, -1.5) {Unlabeled data\\ $(\#x_1, \dots, \#x_m)$};
        \node[emptynode] (predictedlabel) at (6.5, 1.2) {Predicted label?\\ or\\ Human labeled?};
        \node[emptynode, text width=9em] (labeleddata) at (4.5, -1.5) {Labeled data\\ $(\#x_1, \#y_1), \dots, (\#x_n, \#y_n)$};
        
        \draw[->] (unlabeleddata.north) -- (predictor.south) node[pos=0.5, xshift=-1em]{$\#x_i$};
        \draw[->] (predictor.east) -- (xypairs.west) node[pos=0.5, yshift=1.5ex]{$\hat{\#y}_i$};
        \draw[->,dashed] ([xshift=2em]xypairs.north west) -- ([xshift=-1em]judge.south);
        \draw[->,dashed] ([xshift=-2em]xypairs.north east) -- ([xshift=1em]judge.south);
        \draw[->] (judge.east) -- (predictedlabel.west);
        \draw[->] ([xshift=-1em]labeleddata.north) -- ([xshift=2.35em]xypairs.south);
        
        

        \draw[-,dashed,gray] (8,-1.5) -- (8,2);

        \node[redsquarednode] (generator) at (13.5,0) {Generator $G$};
        \node[emptynode] (randomnoise) at (13.5, -1) {Random noise};
        \draw[->] (randomnoise.north) -- (generator.south);
        \node[bluesquarednode] (discriminator) at (10,0) {Discriminator $D$};
        \node[emptynode] (realsamples) at (10, -1) {Real samples $(\#x_1, \dots, \#x_n)$};
        \draw[->] (realsamples.north) -- (discriminator.south);
        \node[emptynode] (realorfake) at (10, 1.25) {Real?\\ or\\ Fake?};
        \draw[->] (discriminator.north) -- (realorfake.south);
        \draw[->] (generator.west) -- (discriminator.east) node[pos=0.5, yshift=1.5ex] {$\hat{\#x_i}$};

    \end{tikzpicture}
    \caption{Left: DAN framework; Right: GAN framework.} \label{fig:DAN} 
\end{figure*}

DAN~\cite{Santos-2017} are adversarial networks framework, that uses only discriminators. The authors propose a DAN formulation suitable for semi-supervised learning. However, we believe that other formulations suitable for the fully supervised learning are also possible. The original DAN use two discriminators: the predictor $P$, and the judge $J$. While $P$ receives a data point $\#x$ on input and outputs a prediction $p(\#x)$, $J$ receives a data point $\#x$ together with a label $\#y$
, and produces a single scalar $J(\#x, \#y)$ representing the probability that $\#x$, $\#y$ came from the labeled training data, rather than being predicted by $P$. See Figure~\ref{fig:DAN} for an illustration of the DAN framework. Note the similarity between DAN and CGAN--- while in CGAN the idea is to generate $\#x$ conditioned on $\#y$, in DAN the goal is to predict $\#y$ conditioned on $\#x$. The cost function for DAN looks as follows:
\begin{eqnarray}
	\min_{P}\max_{J} V(J, P) \!\!\!\! &=& \!\!\!\! \mathbb{E}_{\#x, \#y \sim p_{\mathrm{data}}(\#x, \#y)} \big[ \log J(\#x, \#y) \big] \nonumber \\
     \!\!\!\! & & + \mathbb{E}_{\#x \sim p_{\mathrm{data}}(\#x)} \big[ \log (1 - J(\#x, P(\#x))) \big] \;. \nonumber \\ 
\end{eqnarray}

The important characteristic of this DAN formulation is that $P$ does not make use of labels, so the semi-supervised learning is pretty straightforward within this framework. Even more importantly, there is no need for specification of a loss function for the predictor, it is learned implicitly by the judge. 

Another mention of an implicit loss function learning is in~\cite{Isola-CVPR-2017}, where the model learned to do image-to-image mapping without explicit specification of the loss function. A connection of GAN-based loss function  learning for the generative model, and cost function learning in reinforcement learning (aka inverse reinforcement learning) is presented in~\cite{Finn-NIPS_WAT-2016}. In contrast, DAN concentrates on learning loss functions for discriminative models.

\subsection{Adversarial training/testing}


Adversarial examples were first introduced as attacks to weaken the performance of CNN by addition of noise~\cite{nguyen2015deep}. This led to adversarial training~\cite{Goodfellow-ICLR-2015} where adversarial examples were added to training to make the model more robust. Follow-up work on this eventually led to GAN model. The topic of adversarial training can also be interpreted as loss function learning, i.e. we can use adversarial loss for improving the final classifier robustness. A nice example of adversarial training is EL-GAN~\cite{Ghafoorian-ECCVw-18}, where a GAN framework is used for loss embedding, by which the problem of ill-posed formulation of some tasks is mitigated. Since there are very stringent requirements on safety in AD, the adversarial examples generation might be used as a tool for testing corner cases and robustness. An automated testing mechanism for autonomous driving using deep learning were provided in~\cite{tian2018deeptest}, \cite{pei2017deepxplore} but they do not leverage GAN.





%% file: include/results.tex
\section{OUR RESULTS} 







In this section, we would like to present some of our results from autonomous driving GAN application. To be more specific, we present the results on the soiling and adverse weather classification/enhancement. 

The problem of soling and adverse weather classification consist of the recognition of image deterioration and its possible enhancement. The image deterioration by soiling and adverse weather is caused either by presence of some ``soiling categories'' (e.g. splashes of mud, rain drops, freeze, dust, etc.) on the camera lens, or by adverse weather conditions (e.g. heavy rain, snow, blizzards, etc.). The reasons for dealing with this problem are mainly two-fold: 1) by recognition of soiling type on the camera lens, we can leverage this information to initiate the cleaning system for the lens; 2) we can leverage the information to enhance the image quality (we call this ``desoiling'' in general, it consist of effects such as image ``dehazing'', ``deraining'', etc.). While the reasons for doing 1) are obvious, we can use 2) for example to improve the quality of the recognition pipeline. 

\begin{figure*}[tb]
    \centering
    \includegraphics[width=0.99\linewidth]{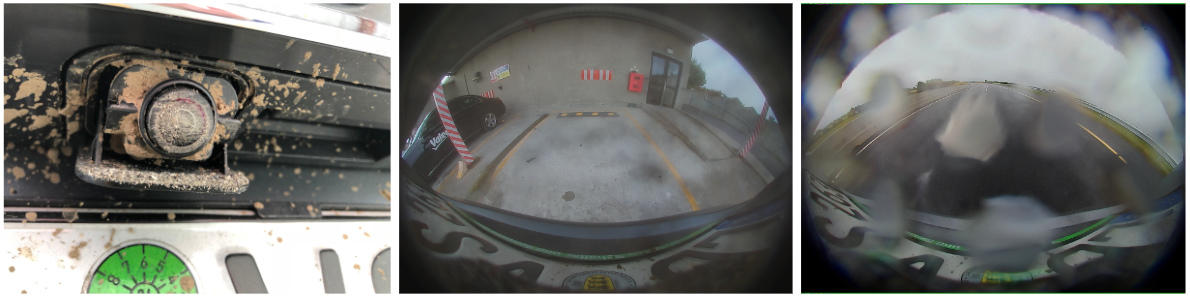}
    \caption{From left to right: a) soiled camera lens mounted to the car body; b) the image quality of the soiled camera from the previous image; c) an example of image soiled by a heavy rain.}
    \label{fig:saw_definition}
\end{figure*}

We decided to use GAN for this problem generally because, as the reader can imagine, obtaining the relevant data is both very problematic and expensive (just imagine that someone has to annotate manually each rain drop on the camera lens during the heavy rain conditions). In Figure~\ref{fig:saw_definition}, we show how the mud splashed on camera can impact the image quality as well as an example of how the image taken during heavy rain looks like. Another reason is the semi-supervised learning potential, which can be easily achieved by GAN. 

\begin{figure*}
    \centering
    \includegraphics[width=0.99\linewidth]{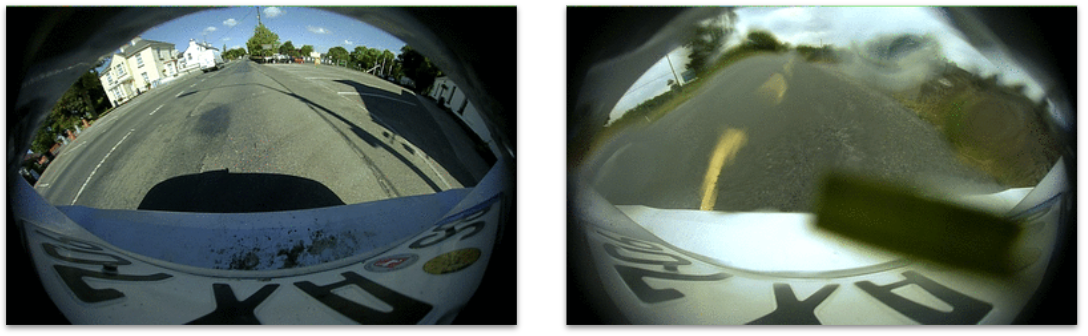}
    \caption{The examples of the ``clean'' and ``soiled'' categories for our CycleGAN experiment.}
    \label{fig:clean_soiled}
\end{figure*}

\begin{figure*}
    \centering
    \includegraphics[width=0.99\linewidth]{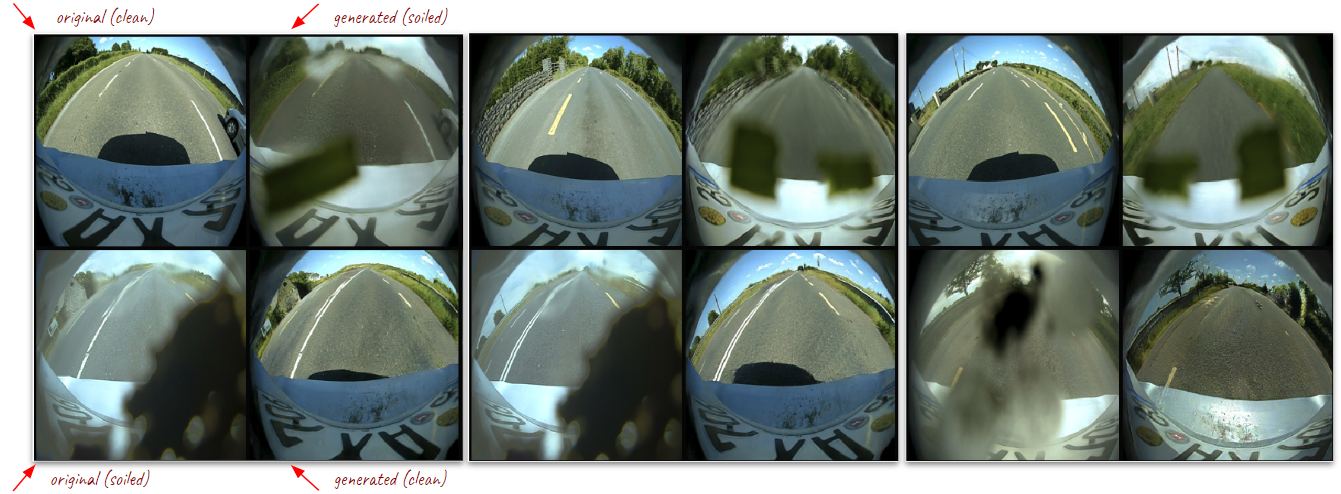}
    \includegraphics[width=0.99\linewidth]{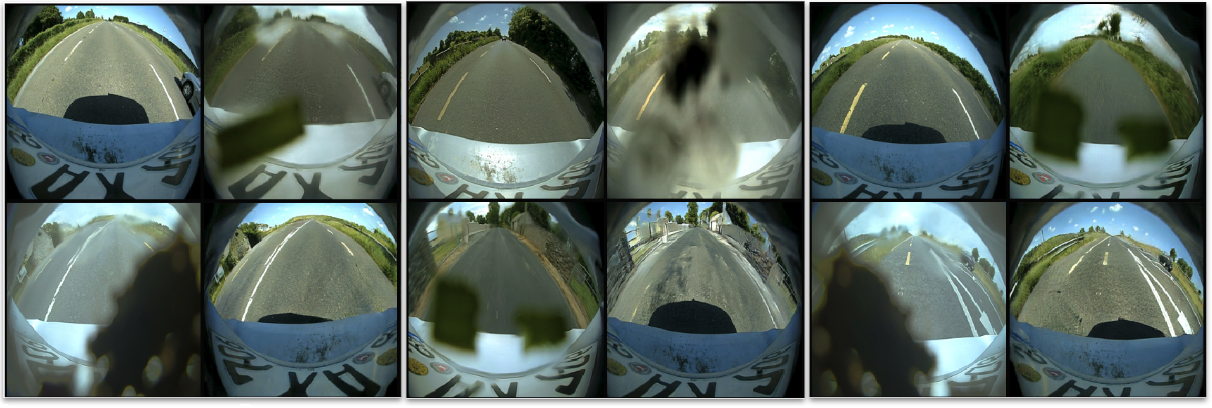}
    \caption{The results obtained from the CycleGAN proof of concept for the problem of soiling and adverse weather classification. Note the legend on the first image, which is self-explanatory.}
    \label{fig:cycleGAN_results}
\end{figure*}

We started with a proof of concept experiment--- we sorted our images to two categories, namely ``clean'' category, which consists of images that are not affected by soiling, and ``soiled'' category, consisting of images deteriorated by the presence of soiling. In Figure~\ref{fig:clean_soiled}, we show the representatives of both classes. This allowed us to try using the CycleGAN~\cite{Zhu-ICCV-2017}. We were happy to see, that just after a few epochs the generator started to correctly recognize which parts of the image are soiled. In the end of the training, we got a generator which is capable of ``desoiling'' the image as well as a generator which can introduce some soiling to the image, see Figure~\ref{fig:cycleGAN_results} for some examples. Note, that due to the relatively small dataset used for this experiment, the ``desoiling'' generator learned to introduce shadow of the car body to the image. This is because the vast majority of images in the ``clean'' category contained shadow of the car body. On the other hand, the ``soiling'' generator learned that the weather was usually cloudy on our images from the ``soiled'' category. 

The CycleGAN experiment encouraged us in the presumption that GAN could be a nice solution for our ptoblem. It also started several hypotheses, such as it might be eventually possible to get semantic segmentation of the soiling without having such explicit annotations, which would be very tedious and expensive to obtain. Another one is that by learning a high quality generators which introduce soiling in the image, we might be able to use them for the advanced data augmentation and create much bigger annotation using all data form various project, which are intentionally ``clean''. 

\begin{figure*}
    \centering
    \includegraphics[width=0.99\linewidth]{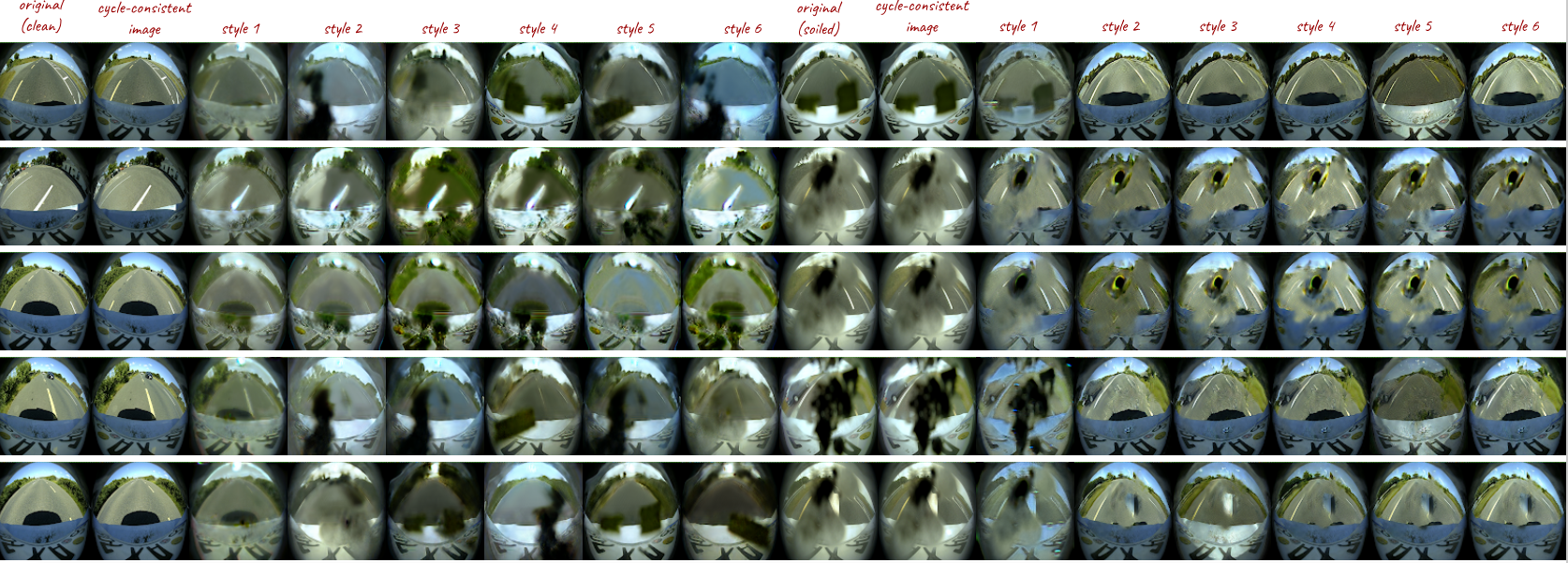}
    \caption{The results from the MUNIT experiment. Note that there are frequent artifacts present. However, it seems that sometimes the generated clean image did not contain the shadow of the body car, and the soiling generator is capable to introduce high variety of soiling categories.}
    \label{fig:munit_results}
\end{figure*}

In the latter direction, the experiment contains an important flaw--- it is not able to produce soling with variable output. Becuase of that, we tried another experiment with the MUNIT~\cite{huang2018multimodal} approach. The main motivation for switching to MUNIT was its ability to split content from the style, which would help our intention to possess the control over generated images and therefore ease the further classifiers training. Our proof of concept for \mbox{MUNIT} failed due to the lack of high quality data. The resulting images from the generator contained a lot of artifacts and the control over the generated classes was not to our satisfaction. However, we still think that it is a promising branch to explore further. We depict our results from MUNIT experiment in Figure~\ref{fig:munit_results}.

We believe that thanks to GAN, the problem of soiling and adverse weather classification/enhancement is possible to be solved even with a minimalist annotation efforts.
We continue our research in the soiling and adverse weather classification/enhancement problematic with a lots of further research direction specified.

%% file: include/discussion.tex
\section{DISCUSSION} \label{sec:challenges}

In this section, we would like to discuss the main challenges of GAN. Namely, we discuss the problem of quantitative evaluation, adversarial attacks with respect to safety, or optimization stability.


\subsection{Quantitative Evaluation}
One of the critical challenges for GAN is their quantitative evaluation. The classic approach for generative models evaluation is based on the model likelihood. However, this approach is is usually intractable~\cite{huszar2015not}. The Inception Score (IS)~\cite{salimans2016improved}, on the other hand, provides a way to quantitatively evaluate the quality and diversity of the generated samples, where conditional label distribution of samples containing meaningful objects should have low entropy and the variability of the samples should be high. Using the generic Inception Net~\cite{szegedy2017inception} trained on ImageNet~\cite{deng2009imagenet}, it was found that IS is well correlated with scores from human annotators. However, IS is found to be insensitive to the prior distribution over labels.
Fr\'{e}chet Inception Distance (FID)~\cite{heusel2017gans} provides another approach for quantifying the quality of generated samples, where the samples are embedded into a feature space given by a specific layer of the Inception Net. Then these are modeled as a continuous multivariate Gaussian distribution. Finally, to quantify the quality of samples, the mean and covariance is estimated for the generated and the real data and the Fr\'{e}chet distance is evaluated. FID score showed to be consistent with human judgment. Unlike IS, FID can detect intra-class mode dropping. IS mainly captures precision where FID captures both precision and recall.  


\subsection{Adversarial examples and Safety}
Adversarial examples are inputs to machine learning models that have been intentionally modified in a way to fool the model. These modifications are, usually, not even noticed by a human observer, yet the classifier still makes wrong classifications. Moreover, adversarial examples can be used to perform attacks on machine learning systems even in a physical world. It was shown in~\cite{kurakin2016adversarial} that machine learning systems are vulnerable to adversarial examples in physical world scenarios. Most of adversarial example attacks require knowledge of either the model or its training data. However, \cite{papernot2017practical} introduced a practical demonstration of an attacker with no such knowledge known as black box attacks. For the autonomous driving, attackers can target autonomous vehicles by using stickers to create an adversarial sign that the vehicle would interpret as another other sign leading to performing unwanted or even dangerous behaviour. In the Universal Adversarial Training~\cite{shafahi2018universal}, instead of adding tailored perturbations to an image, an update can be added to any image in a broad class of images, while changing the predicted classes. One example for defensive approaches is the Defensive Distillation mechanism~\cite{papernot2016distillation} that trains a model whose surface is smoothed in the directions an attacker will typically try to exploit, making it difficult to discover adversarial examples. Furthermore, Reinforcement Learning (RL) agents can also be manipulated by adversarial examples which causes a degraded or even a dangerous policy. Adversarial examples represent a concrete problem in AD safety and designing methods for preventing adversarial examples is an active area of research.\\

\subsection{Optimization Stability}

GAN training requires to find a Nash equilibrium~\cite{Nash1951} of non-convex non-cooperative game with high-dimensional parameters. The training typically consist of some form of gradient descent. Moreover, the optimization is, due to convergence issues, designed to minimize some loss function, rather than to find the Nash equilibrium. In~\cite{salimans2016improved}, the authors use the heuristic understanding of the non-convergence problem to introduce several techniques for its improving, such as feature matching trying to prevent overtraining of the discriminator, minibatch discrimination, which, by allowing the discriminator to see several data samples in combination, prevents generator from collapsing to always generate the same sample, and many other ``hacks'' trying to overcome the identified problems. 

The authors of~\cite{pmlr-v70-arjovsky17a} tried to approach the known convergence and stability problems of GAN training differently. They described the source cause of these problems by identifying the Kullback-Leibler (KL) divergence minimization task hidden in the probability distribution learning. The problem arises when dealing with distributions supported by low-dimensional manifolds, because for these the KL is either not defined or simply infinite. As a solution to overcome this problem, the authors propose to use a different distance function, which consists of the earth-mover, or Wasserstein, distance. In experimental evaluation, they show that the desirable stabilization effect is achieved. The algorithm proposed in~\cite{pmlr-v70-arjovsky17a} have one flaw, which is the clipping of weights to enforce $1$-Lipshitzness required in the optimization theory of WGAN. This flaw was corrected by~\cite{Gulrajani-NIPS-2017}, where the gradient penalty was introduced. 


%% file: include/conclusions.tex
\section{CONCLUSIONS} \label{sec:conclusions}

We compiled a detailed overview of GAN models and provide a taxonomic survey of various applications of GAN in autonomous driving.  GAN have a potential for high impact for autonomous driving applications but there is slow progress in this area. There are plenty of other applications beyond the standard image translation application. We also discussed the main challenges and open problems which have to be resolved in order for it to be more practically used. We hope that this paper encourages further research in applying GAN for autonomous driving applications.

%% file: include/aboutauthors.tex
\begin{biography}

\vskip2ex

\noindent \textbf{Michal U\v{r}i\v{c}\'{a}\v{r}} received MSc. degree (major in computer graphics and minor in computer vision) from Czech Technical University in Prague in 2011, and Ph.D. from the same university (in 2018) while working in the machine learning group of the Center for Machine Perception under the  supervision of Dr. Vojt\v{e}ch Franc. Since 2016, he is working as a researcher in Valeo R\&D, Prague. His main research focus is machine learning, with applications in computer vision. He received the best paper prize at VISAPP in 2012 and the 3rd place at Looking at People Challenge: Age estimation track, organized at CVPR 2016.\\

\noindent \textbf{Pavel K\v{r}\'{i}\v{z}ek} received M.Sc. in Control Engineering from the Czech Technical University in Prague in 2003, and Ph.D. in Bio-Cybernetics and Artificial Intelligence from the same university in 2008 at the Center for Machine Perception. During Ph.D., he spent two years at the Center for Vision, Speech and Signal Processing at University of Surrey, UK. Over five years he was working as a Postdoc on several projects focusing on super-resolution imaging in florescence microscopy at the Charles University in Prague. Since 2014, he is a researcher in Valeo R\&D, Prague. He has over 16 years of experience in image processing, computer vision, and machine learning applications including 4 years of experience in industrial automotive systems. He is an author of 30 peer reviewed journal and conference papers and 4 patents. \\

\noindent \textbf{David Hurych} received his Ph.D. in Bio-Cybernetics and Artificial Intelligence from Czech Technical University in Prague in 2014. He received his M.Sc. from Computer Science from VSB - Technical University of Ostrava in 2007. Since 2014 he is a researcher and team leader in Valeo R\&D, Prague and was awarded machine learning expert price. His main research focus is machine learning with applications in computer vision. He is an author of 10 conference and journal papers (including one TPAMI) and 4 awarded patents. He received the best student paper award at VISAPP 2011. He has 11 years of machine learning and computer vision experience including 5 years of experience in automotive industry and autonomous driving. \\

\noindent \textbf{Ibrahim Sobh} received Ph.D in Deep Reinforcement Learning for fast learning agents acting in 3D environments directly from high-dimensional sensory inputs, form Cairo University Faculty of Engineering, in 2017. He received his B.Sc., in 1997, and M.Sc., in 2009, from the same University. His M.Sc. thesis is in the field of Machine Learning and applied on automatic documents summarization. Ibrahim has participated in several related national and international mega projects, conferences and summits. Ibrahim's publications including international journals and conference papers are mainly in the machine and deep learning fields for Natural language processing, Speech processing, Computer vision and autonomous driving. \\

\noindent \textbf{Senthil Yogamani} is a computer vision architect and technical leader at Valeo Vision systems. He is currently focused on research and design of the overall computer vision algorithm architecture for surround-view camera visual perception in autonomous driving systems. He has over 13 years of experience in computer vision and machine learning including 10 years of experience in industrial automotive systems. He is an author of 50 peer reviewed publications and 33 patents. He serves in the editorial board of various leading IEEE automotive conferences including ITSC and ICVES and advisory board of various industry consortia including Khronos, Cognitive Vehicles and IS Auto. He is a recipient of best associate editor award at ITSC 2015 and best paper award at ITST 2012. \\

\noindent \textbf{Patrick Denny} is a Senior Research Engineer and Senior Expert at Valeo Vision Systems, where he has been a Technical Lead and internal consultant on RF, camera and imaging system developments for the last 17 years for the leading automotive marques. He is Adjunct Professor of Automotive Electronics in NUI Galway. He received his B.Sc. in Experimental Physics and Mathematics from NUI Maynooth in 1993, an M.Sc. in Pure Mathematics from NUI Galway in 1994 and a Ph.D in Physics in 2000 from NUI Galway and a Professional Diploma in Data Analytics from University College Dublin in 2016. He has 119 patent applications in 60 patent families and over 30 publications in refereed journals, conference proceedings and books. He co-founded and chairs and is on the steering committee of the AutoSens and Electronic Imaging's Autonomous Vehicles and Machines conferences and co-founded and chaired IS Auto. He is also a co-founder and board member of the IEEE P2020 Automotive Imaging Standards body. \\

\end{biography}